\documentclass[letterpaper, 10 pt, conference]{ieeeconf}  

\usepackage{comment}
\usepackage{graphicx}
\usepackage{hyperref}
\usepackage{balance}
\usepackage{cite}
\usepackage{amssymb}
\usepackage{amsmath}
\usepackage{mathtools}
\usepackage{mathastext}
\DeclareMathOperator*{\argmin}{arg\,min}

\usepackage[linesnumbered,ruled,noend,vlined]{algorithm2e}
\usepackage{algs}
\SetArgSty{textup}

\SetCommentSty{mycommfont}
\bstctlcite{IEEEexample:BSTcontrol}
\usepackage[dvipsnames]{xcolor}

\IEEEoverridecommandlockouts                              

\title{\LARGE \bf \textit{Strobe}: An Acceleration Meta-algorithm for Optimizing Robot Paths using Concurrent Interleaved Sub-Epoch Pods}

\author{Daniel Rakita$^{1}$, Bilge Mutlu, Michael Gleicher

\thanks{$^{1}$Authors are with the Department of Computer Sciences, University of Wisconsin--Madison, Madison 53706, USA
{\tt\small [rakita|bilge|gleicher]@cs.wisc.edu}}%
\thanks{This work was supported by a Microsoft Research PhD Fellowship, National Science Foundation award 1830242, and a NASA University Leadership Initiative (ULI) grant awarded to the UW-Madison and The Boeing Company (Cooperative Agreement \# 80NSSC19M0124).}%
}

\begin{document}

\maketitle
\thispagestyle{empty}
\pagestyle{empty}



\begin{abstract}
In this paper, we present a meta-algorithm intended to accelerate many existing path optimization algorithms.  The central idea of our work is to strategically break up a waypoint path into consecutive groupings called ``pods,'' then optimize over various pods concurrently using parallel processing.  Each pod is assigned a color, either blue or red, and the path is divided in such a way that adjacent pods of the same color have an appropriate buffer of the opposite color between them, reducing the risk of interference between concurrent computations.  We present a path splitting algorithm to create blue and red pod groupings and detail steps for a meta-algorithm that optimizes over these pods in parallel.  We assessed how our method works on a testbed of simulated path optimization scenarios using various optimization tasks and characterize how it scales with additional threads.  We also compared our meta-algorithm on these tasks to other parallelization schemes.  Our results show that our method more effectively utilizes concurrency compared to the alternatives, both in terms of speed and optimization quality.    
\end{abstract} 

\section{Introduction}
\label{sec:intro}

Many sub-problems in robotics require optimizing over \textit{paths}, \textit{i.e.,} ordered sequences of states within a configuration space \cite{lavalle2006planning}.  For example, a hospital delivery robot may optimize its navigation trajectory to maximize distance from oncoming people in a hallway, or a home-care robot manipulator may optimize its joint-position path to move smoothly while maintaining an upright end-effector pose through a motion to avoid spilling a glass of water.  A common strategy for optimizing over robot paths is to (1) discretize the path into a fixed set of waypoints connected by splines (often linear splines); and (2) use an optimization method to minimize some objective function relating to the set of waypoints subject to any constraints, starting from some initial condition.  The objective function is often highly non-linear, especially in robotics problems, which adds to the computational complexity and typically precludes provably global optimal solutions.   

While the path optimization strategy above is often successful for at least finding feasible, locally optimal solutions, especially when starting from a good initial condition, there are challenging trade-offs in practice.  In particular, using a large number of waypoints is often desirable as it gives the optimizer more representational power to shape the path based on the specified objective function and constraints.  However, the more waypoints are used, the longer it takes the optimizer to converge, often taking minutes or hours to find a solution in the case of dense paths, especially if numerous initial conditions must be tried.  These challenges significantly hinder path optimization methods from being used in important real-time or semi-online settings, such as shared control, teaching by demonstration, or inverse reinforcement learning, especially if many waypoints are needed to sufficiently parameterize the path.    

In this paper, we present a meta-algorithm, called \textit{Strobe}, intended to accelerate many path optimization algorithms.  The central idea of our work is to strategically split the waypoint path into consecutive groupings called ``pods,'' then optimize over various pods concurrently using parallel processing.  Each pod is assigned a color, either blue or red, and the path is divided in such a way that adjacent pods of the same color have a buffer of the opposite color between them.  Thus, each pod can reasonably assume that its surrounding neighborhood on the path is somewhat constant with respect to its own optimization, and each thread can focus on optimizing a small segment of the path with low risk of interfering with another thread's optimization.  The method iteratively interleaves optimizing over blue and red pod groups until the process either converges or reaches a maximum number of epochs.  



\begin{figure}[t!]
	\includegraphics[width=\columnwidth]{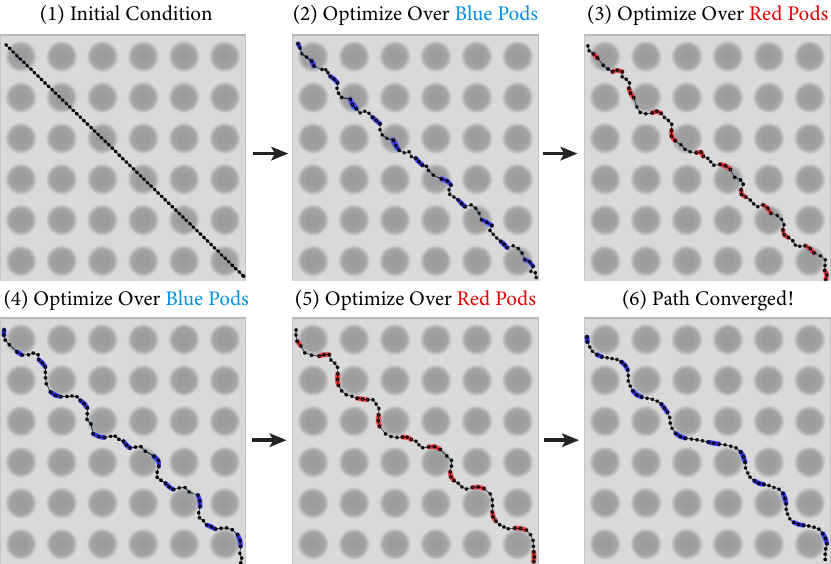}
	\caption{We present a meta-algorithm, called \textit{Strobe}, intended to accelerate path optimization algorithms.  This figure illustrates our meta-algorithm optimizing a path to avoid areas of high cost (circles) in the image.  Our method splits the path into blue and red consecutive subsets called ``pods,'' and alternates optimizing over these groups until the path converges.}
	\label{fig:teaser}
	\vspace{3pt}
\end{figure}

Our work offers three main contributions: (1) presenting a path splitting algorithm that attempts to maximize the separation between adjacent pods of the same color while minimizing the number of waypoints per pod given any number of available threads; (2) formulating a meta-algorithm that optimizes over these pods in parallel; and (3) providing open-source code for an implementation of our method.\footnote{\href{https://github.com/uwgraphics/lynx}{https://github.com/uwgraphics/lynx}}  

We assessed the efficacy of our method by running a testbed of experiments involving two-dimensional path optimization and higher dimensional robot path optimization tasks (\S\ref{sec:evaluation}).  We assessed how our method works with various optimization algorithms and characterize how it scales when adding more threads.  We also compared our meta-algorithm to other parallelization schemes on these tasks.  Our results show that our method more effectively utilizes concurrency compared to the alternatives, both in terms of speed and optimization quality.  Interestingly, our method often converges on higher quality paths than even its single-threaded counterparts.  We conclude with an overall discussion about the implications of our work based on our results (\S\ref{sec:discussion}).   






\section{Related Works}
\label{sec:related_works}

\textit{Motion Planning}---Robot motion planning is the process of finding collision-free, feasible paths from a start state to a goal state in configuration space \cite{lavalle2006planning}.  Sampling-based motion planners often use random samples to bootstrap a search strategy and build a graph structure from start to goal \cite{lavalle1998rapidly, kavraki1996probabilistic, hsu1997path, kuffner2000rrt}.  Solutions found from these methods often have many extraneous and unnecessary motions, thus post-processing of the motions is common.  We believe our method could help at this post-processing phase by quickly optimizing over an already feasible path, \textit{e.g.,}, smooth velocities to shorten the path or adjust the motion such that a robot has a specified end-effector orientation throughout.  Some sampling-based planners can accommodate limited objectives, such as achieving a shortest path in the limit \cite{karaman2011sampling, gammell2014informed}, though this additional objective comes at the cost of more computation time and cannot accommodate arbitrary objectives.   

\textit{Trajectory Optimization}---Path optimization has many similarities with the area of \textit{trajectory optimization}, an approach used to optimize the motion of a body over time to match desired motion qualities (see Betts \cite{betts1998survey} for a review).  In the context of robotics, trajectory optimization often involves optimizing a dynamical system that includes both state and control variables over time \cite{toussaint2009robot, posa2014direct}.  For example, much work on quadruped and humanoid robots has been done on optimizing over the kinematic states, dynamics states, and joint torque control inputs of the robot to create robust and fluid locomotion trajectories \cite{kuindersma2016optimization, dai2014whole}.  In this work, we only focus on optimizing over states along a path and not over control variables and time as well.  However, we believe trajectory optimization methods, especially those that use direct collocation techniques, may benefit from the acceleration ideas presented in our work.  We plan to explore these possibilities in future work.   

Some trajectory optimization approaches in robotics are used primarily for kinematic, motion planning-based tasks \cite{ratliff2009chomp, kalakrishnan2011stomp, park2012itomp, schulman2013finding, marinho2016functional}.  Many of the gradient and non-gradient based optimization algorithms we test our method on in this work share similarities with the optimization procedures underlying these proposed approaches, so believe our method would be compatible with these approaches as well.

\textit{Optimization in Parallel}---Our work draws on prior work that uses concurrency to accelerate optimization and planning procedures.  Betts et al. proposed a scheme for performing trajectory optimization using a parallel shooting method \cite{betts1991trajectory}.  Paine et al. presented a method that parallelized the ITOMP trajectory optimization algorithm where many initial conditions were tried at once \cite{park2013real}.  We note that our method differs from a multiple-restart parallel algorithm, as our goal is to accelerate a single optimization path instance instead of trying multiple at once.  Also, stochastic gradient descent (SGD) is a popular technique for optimizing high dimensional functions in parallel, particularly popularized by the training of neural networks \cite{paine2013gpu}.  Path optimization differs from the problems that SGD is commonly used on as it commonly considers derivatives and continutity between points rather than treating parts of the optimization vector as independent sub-problems.  We compare our method to a generalized SGD strategy in our evaluation.

\section{Technical Overview}
\label{sec:technical_overview}

In this section, we provide preliminaries about notation and the problem we are attempting to accelerate, and we overview the main technical premise of our method.  

\subsection{Path Optimization Problem Statement}
\label{sec:preliminaries}
A \textit{path}, which we will denote as $\Gamma$, can be thought of as a function that maps some parameter, $\textit{u} \in [0,1]$, to some state, $\mathbf{q}$, within an $\textit{n}$-dimensional configuration space, $\mathbf{q} \in \chi$.  We will denote the space of all possible paths in $\chi$ as $\Xi$.  From these definitions, the path optimization problem is as follows:


\footnotesize
\begin{equation}
\label{eq:problem_statement_continuous}
\begin{gathered}
\Gamma^* = \argmin_{\Gamma \in \Xi} \mathbf{f}(\Gamma)\\
\mathbf{f}(\Gamma) =\mathbf{g}(\Gamma(0), \Gamma(1)) + \int_0^1 \mathbf{h}(\Gamma(\textit{u}), \Gamma'(\textit{u}), \Gamma''(\textit{u}), ...) \ \textit{du} \\
s.t. \ \ \mathbf{c}_{be}(\Gamma(0), \Gamma(1)) = \mathbf{0}, \ \ \mathbf{c}_{bi}(\Gamma(0), \Gamma(1)) < \mathbf{0} \\
\ \ \mathbf{c}_{e}(\Gamma(\textit{u}), \Gamma'(\textit{u}), \Gamma''(\textit{u}), ...) = \mathbf{0} \ \forall \textit{u} \in [0,1], \\ 
\ \ \mathbf{c}_i(\Gamma(\textit{u}), \Gamma'(\textit{u}), \Gamma''(\textit{u}), ...) < \mathbf{0} \ \forall \textit{u} \in [0,1]
\end{gathered}
\end{equation}
\normalsize

Here, $\mathbf{g}$ is a function that characterizes the quality of the path boundary points, $\mathbf{h}$ is a function that characterizes the quality of an interior point on the path and any of its derivatives at that point, $\mathbf{c}_{be}$ and $\mathbf{c}_{bi}$ are sets of equality and inequality constraints, respectively, that dictate whether particular boundary points are feasible, and $\mathbf{c}_{e}$ and $\mathbf{c}_{i}$ are sets of equality and inequality constraints, respectively, that dictate whether interior points along the path and any of their respective derivatives are feasible.  Any of the functions in Equation \ref{eq:problem_statement_continuous} can be non-linear, which is often the case for robotics-related path optimization problems. 

\subsection{Path Optimization Discretization}
\label{sec:path_optimization_discretization}
While it may be possible to analytically solve some problems using Equation \ref{eq:problem_statement_continuous}, \textit{e.g.}, using calculus of variations \cite{gelfand2000calculus}, it is common to solve difficult, real-world problems by discretizing the path optimization problem and using numerical methods over a finite set of variables.  Formally, we consider a path $\Gamma$ to be parameterized by a set of $M$ ordered waypoint states, $\mathcal{W}$.  We will use $\mathcal{W}[\textit{i}]$ to refer to the $\textit{i}$-th waypoint in the set $\mathcal{W}$.  If all waypoint states are concatenated together into a single vector, we will denote this as $\mathbf{X}$ and call it the \textit{full path vector}, \textit{i.e.,} $\mathbf{X} \equiv [\mathcal{W}[0]^\top \ \mathcal{W}[1]^\top  \cdots \ \mathcal{W}[M]^\top ]^\top \in \mathbb{R}^{M * n}$.  The waypoints in $\mathcal{W}$ can always be converted to a continuous path $\Gamma$ by treating the waypoints as knots along splines.  For instance, it is standard to use linear splines, meaning the resulting continuous path would be linear interpolation between all waypoints from $\mathcal{W}[0]$ to $\mathcal{W}[M]$.



Using the notation above, a discrete version of Equation \ref{eq:problem_statement_continuous} can be formulated as follows:

\footnotesize
\begin{equation}
\label{eq:problem_statement_discrete}
\begin{gathered}
\mathbf{X}^* = \argmin_{\mathbf{X} \in \mathbb{R}^{M * n}} \mathbf{f}(\mathbf{X}) \\
\mathbf{f}(\mathbf{X}) = \mathbf{g}(\mathcal{W}[0], \mathcal{W}[M]) + \sum_{\textit{i}=0}^M \ \mathbf{h}( \mathcal{W}[\textit{i}], \mathcal{W}'[\textit{i}], \mathcal{W}''[\textit{i}], ... ) \\
s.t. \ \mathbf{c}_{be}(\mathcal{W}[0], \mathcal{W}[M]) = \mathbf{0}, \ \ \mathbf{c}_{bi}(\mathcal{W}[0], \mathcal{W}[M]) < \mathbf{0}, \\
\ \ \mathbf{c}_{e}(\mathcal{W}[\textit{i}], \mathcal{W}'[\textit{i}], \mathcal{W}''[\textit{i}], ...) = \mathbf{0} \ \forall \textit{\textit{i}} \in \{0,1,...,M\}, \\ 
\ \ \mathbf{c}_i(\mathcal{W}[\textit{i}], \mathcal{W}'[\textit{i}], \mathcal{W}''[\textit{i}], ...) < \mathbf{0} \ \forall \textit{\textit{i}} \in \{0,1,...,M\}
\end{gathered}
\end{equation}
\normalsize

Here, the derivatives at the waypoints $\mathcal{W}[i]$ are typically approximated using the neighboring predecessors and successors, \textit{e.g.,} using finite differencing.  

Our work is focused on accelerating numerical optimization methods used to solve Equation \ref{eq:problem_statement_discrete}.  In the following section, we overview the structure of our meta-algorithm that achieves this acceleration.


\subsection{Strobe Overview}
\label{sec:strobe_overview}
\begin{figure}[t!]
	\includegraphics[width=\columnwidth]{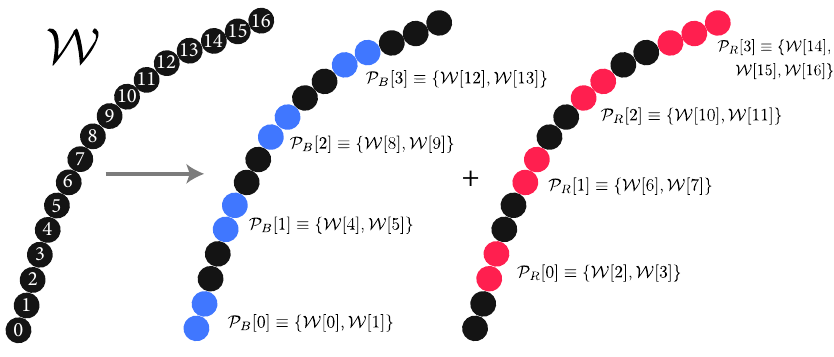}
	\caption{Waypoints along path being divided into blue and red pods.}
	\label{fig:pod_split}
	\vspace{2pt}
\end{figure}

The central idea of our method is to strategically split the set of waypoints $\mathcal{W}$ into smaller consecutive groupings called \textit{pods}, then optimize over various pods concurrently using parallel processing (either multi-threaded CPU or GPU).  A pod is a consecutive subset of waypoints in $\mathcal{W}$ that is assigned a color, either blue or red.  If a pod is blue, it is considered part of the set $\mathcal{P}_B$, and conversely, if a pod is red, it is considered part of the set $\mathcal{P}_R$.  Thus, $\mathcal{P}_B$ and $\mathcal{P}_R$ are sets of consecutive $\mathcal{W}$ subsets.  We will refer to the $\textit{i}$-th set in $\mathcal{P}_B$ or $\mathcal{P}_R$ as $\mathcal{P}_B[\textit{i}]$ and $\mathcal{P}_R[\textit{i}]$, respectively.  An illustration of these sets can be seen in Figure \ref{fig:pod_split}.  It is required that all waypoints in $\mathcal{W}$ belong to exactly one pod.   


Using the definition of pods, the Strobe meta-algorithm follows three steps: (1) Optimize over all blue pods in parallel; (2) Optimize over all red pods in parallel; (3) Check if the whole path has converged subject to some tolerance.  If it has converged, return the path, and if not, go back to step 1.  Here, steps 1 and 2 can accommodate any base optimization algorithm that is compatible with the model's underlying objectives and constraints.  Also, note that all waypoints will be optimized over after each pass through steps 1 and 2.  We refer to one pass through both steps 1 and 2 as an \textit{epoch} and each pass through either blue or red pods (step 1 or 2, respectively) as \textit{sub-epochs}.       

A challenge addressed in this work is how to effectively divide the waypoint path into separate pods.  Our goal is to split the path in such a way that the whole path successfully converges, even with much work being done on separate threads that are not explicitly communicating with one another.  We divide the path based on two heuristics: (1) The number of blue and red pods should be as high as possible, but no more than the number of available threads on the processor.  This will help evenly distribute the work being done at once in steps 1 and 2 above; and (2) Pods should be no less than $\ell$ distance away (in terms of waypoint index distance) of the nearest pod of the same color.  This helps ensure that parallel optimizations do not cross-pollinate and inadvertently form contradicting strategies toward convergence.  It is good to choose $\ell$ based on the number of predecessors and successors used for approximating derivatives at a given waypoint in Equation \ref{eq:problem_statement_discrete}.  In Figure \ref{fig:pod_split}, $\ell = 2$, which is a reasonable value for many applications.







\section{Technical Details}
\label{sec:technical_details}
In this section, we present our path splitting algorithm based on the heuristics above, and provide more details about the Strobe meta-algorithm.  

\subsection{Path Splitting Algorithm}
\label{sec:path_splitting}

\algPathSplitting
\algAddPod
\algStrobe

Our path splitting algorithm can be seen in Algorithms \ref{alg:path_splitting} and \ref{alg:add_pod}.  The algorithm progresses in four steps.  In step 1, the algorithm calculates a low number of waypoints per pod, \textit{wpp\_min}, and a high number of waypoints per pod, \textit{wpp\_max}, based on the number of threads available and $\ell$.  Note that the minimum number of waypoints per pod will never be less than $\ell$ based on the distance constraint, and \textit{wpp\_min} will always be $\textit{wpp\_max}-1$.  The values for \textit{wpp\_min} and \textit{wpp\_max} are selected such that they are the lowest values possible that will be guaranteed to cover all $M$ waypoints given the number of threads available.  This helps achieve heuristic 1 outlined in \S\ref{sec:strobe_overview} above.  In step 2, the algorithm computes how many pods of length \textit{wpp\_min} and \textit{wpp\_max} it will take to reach the total number of waypoints, $|\mathcal{W}|$, denoted as \textit{n\_min} and \textit{n\_max}, respectively.  These values will be exact unless the number of waypoints is too low to utilize all available threads.

In step 3, The algorithm adds \textit{n\_min} pods of length \textit{wpp\_min} and \textit{n\_max} pods of length \textit{wpp\_max} to a holding set $\mathcal{A}$.  If at any point during these pod additions the total number of waypoints is reached (meaning the numbers in step 2 were overestimates), the remainder of the waypoints are either absorbed by the previously added pod or are appended as their own pod of truncated size.  This decision is based on whether the excess nodes are greater than or less than $\ell$ as the method will not allow a pod of size less than $\ell$.  Lastly, step 4 involves the algorithm assigning every other pod in the set $\mathcal{A}$ to be either blue or red, such that they alternate from start to finish.  Alternating colors between pods that are each guaranteed to be no less than length $\ell$ will ensure that heuristic 2 outlined in \S\ref{sec:strobe_overview} will also be achieved.                   





\subsection{Strobe Details}
\label{sec:strobe_details}

A pseudocode view of the Strobe meta-algorithm can be seen in Algorithm \ref{alg:strobe}.  At a high level, the algorithm alternates between optimizing over blue and red pods in parallel over some maximum number of epochs or until the optimization converges subject to some specified tolerance.  The \texttt{optimize} input can be any optimization algorithm that is compatible with the underlying objectives and constraints.   

Because the optimizations per thread are only happening over a subset of the full path on a single pod, each must optimize a modified version of Equation \ref{eq:problem_statement_discrete}:

\footnotesize
\begin{equation}
\label{eq:problem_statement_with_split}
\begin{gathered}
\mathbf{X}[\textit{a}:\textit{b}]^* = \argmin_{\mathbf{X}[\textit{a}:\textit{b}] \in \mathbb{R}^{(\textit{b} - \textit{a} + 1) * n}} \mathbf{f}_s(\mathbf{X}, \textit{a}, \textit{b}) \\
\mathbf{f}_s(\mathbf{X}, \textit{a}, \textit{b}) = \mathbf{g}(\mathcal{W}[0], \mathcal{W}[M]) + \sum_{\textit{i}=\textit{a}}^\textit{b} \ \mathbf{h}( \mathcal{W}[\textit{i}], \mathcal{W}'[\textit{i}], \mathcal{W}''[\textit{i}], ... ) \\
s.t. \ \mathbf{c}_{be}(\mathcal{W}[0], \mathcal{W}[M]) = \mathbf{0}, \ \ \mathbf{c}_{bi}(\mathcal{W}[0], \mathcal{W}[M]) < \mathbf{0}, \\
\ \ \mathbf{c}_{e}(\mathcal{W}[\textit{i}], \mathcal{W}'[\textit{i}], \mathcal{W}''[\textit{i}], ...) = \mathbf{0} \ \forall \textit{\textit{i}} \in \{0,1,...,M\}, \\ 
\ \ \mathbf{c}_i(\mathcal{W}[\textit{i}], \mathcal{W}'[\textit{i}], \mathcal{W}''[\textit{i}], ...) < \mathbf{0} \ \forall \textit{\textit{i}} \in \{0,1,...,M\}
\end{gathered}
\end{equation}
\normalsize 

Here, $\textit{a}$ and $\textit{b}$ are the lower and upper waypoint index bounds of the path being optimized over, respectively.  Note that if $\textit{a} > 0$ and $\textit{b} < M$, the $\mathbf{g}$ objective and $\mathbf{c}_{be}$ and $\mathbf{c}_{bi}$ constraints do not have to be considered in the particular optimization as these terms will be constant with respect to the given path segment.        

The only parts of our meta-algorithm that involve the separate threads writing to a shared data structure occurs at Algorithm \ref{alg:strobe} lines 9 and 14 when the full state vector is updated based on each thread's individual optimization.  This writing process is quite fast and we find that this requires little synchronization overhead in practice.

\section{Evaluation}
\label{sec:evaluation}
In this section, we overview the experiments designed to assess the efficacy of our method. 

\subsection{Implementation Details}
\label{sec:implementation_details}
A prototype implementation of our method was implemented in the Rust programming language.  Parallel operations were run using the Rayon multithreading library\footnote{\href{https://docs.rs/rayon/1.5.0/rayon/}{https://docs.rs/rayon/1.5.0/rayon/}}.  All evaluations were run on a Lenovo Legion laptop with an i7-9750H processor with six physical cores (twelve threads) and 32GB RAM.  


\begin{figure*}[t!]
	\includegraphics[width=\textwidth]{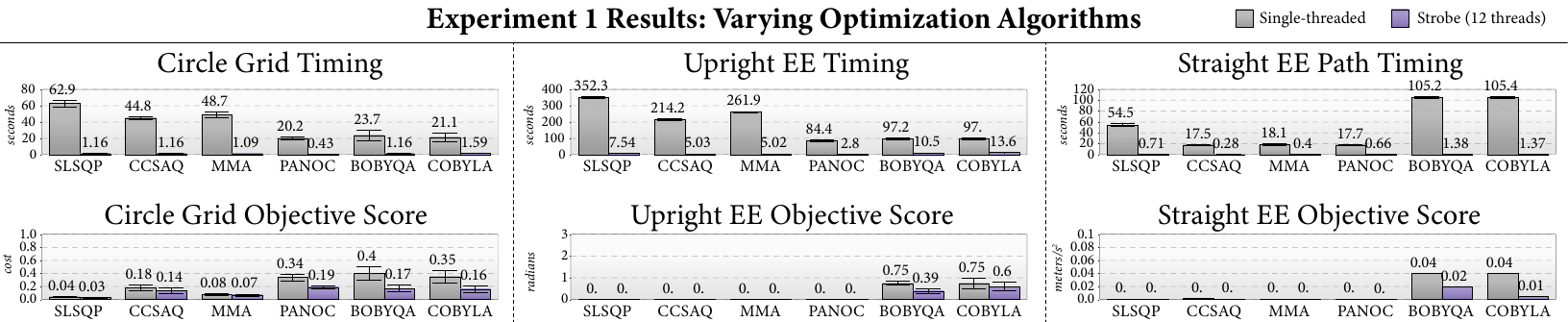}
	\caption{Results from Experiment 1.  Range values denote standard error.  Gray bars on the left show results for the optimization algorithm run on a single thread, while the purple bars on the right show the results for the optimization algorithm run using the Strobe meta-algorithm. }
	\label{fig:experiment1}
	\vspace{2pt}
\end{figure*}

\begin{figure*}[t!]
	\includegraphics[width=\textwidth]{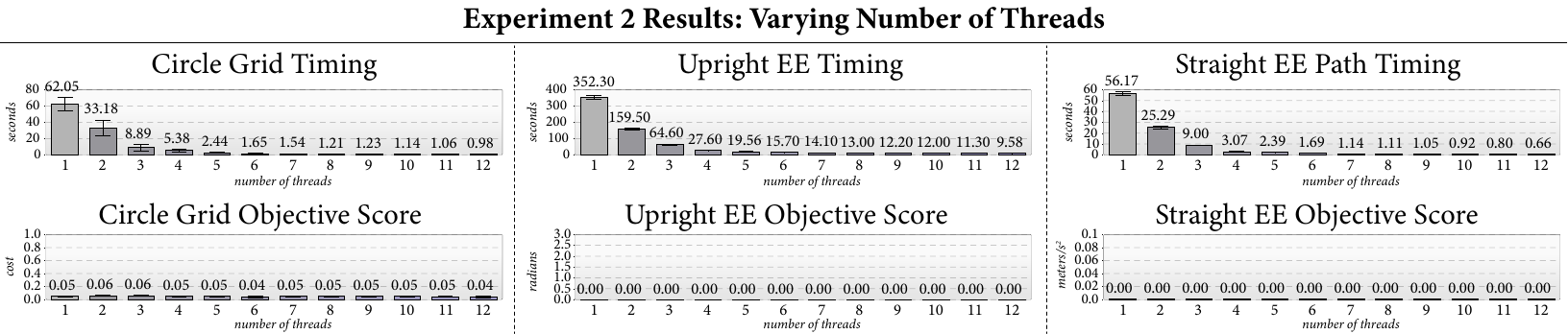}
	\caption{Results from Experiment 2.  Range values denote standard error.}
	\label{fig:experiment2}
	\vspace{-5pt}
\end{figure*}

\subsection{Evaluation Benchmark}
We developed three benchmark scenarios to evaluate our method.  The first scenario is the 2-dimensional \textit{Circle Grid} planning example seen in Figure \ref{fig:teaser}.  Here, the dark circles represent areas of high cost, and the primary objective is to steer the path to lighter areas of lower cost.  The path also has additional objectives to maintain smooth derivatives.  The second scenario, \textit{Upright EE} involves a 7-DOF Sawyer robot optimizing its motion to exhibit an upright end-effector orientation throughout the motion.  The third scenario, \textit{Straight EE Path} involves the robot optimizing its motion to minimize acceleration in end-effector space, thus tracing a straight line with its end-effector translation.  In both of the Sawyer scenarios, the robot has additional objectives to maintain smooth joint velocities, acceleration, and jerks, and also avoid self-collisions.

Every condition tested in our experiments performed these scenarios 100 times with different initial conditions each time.  All initial conditions involved taking a straight line between random points of a fixed distance within the bounds of the planning space, populating a desired number of waypoints along this line, and adding uniformly random noise to each waypoint.  All conditions used identical initial conditions in our experiments to ensure fair comparisons.  We set a maximum time limit of 20 minutes for each task.    

\subsection{Metrics}
We report on two metrics per task: the computation time it took to converge, and some information per task that gives a sense of optimization quality.  For the \textit{Circle Grid} task, our optimization quality metric was the average ``image cost'' of the waypoints along the path, where dark areas of the image have cost 1, and light areas of the image have cost 0.   For the \textit{Upright EE} task, our optimization quality metric was the average rotation error away from the given goal orientation (in radians).  For the \textit{Straight EE Path} task, our optimization quality metric was the average acceleration in end-effector space along the path, where 0 designates an exact line.  These accelerations were approximated by finite differencing over the waypoints along the path.  


\subsection{Experiment 1: Varying Optimization Algorithms}
In Experiment 1, we assessed the performance of our method on six optimization algorithms: COBYLA (constrained optimization by linear approximation) \cite{powell1994direct}, BOBYQA (bound optimization by quadratic approximation) \cite{powell2009bobyqa}, MMA (method of moving asymptotes) \cite{svanberg2002class}, CCSAQ (conservative convex separable approximation) \cite{svanberg2002class}, SLSQP (sequential least-squares quadratic programming) \cite{kraft1988software}, and PANOC \cite{stella2017simple}.  The COBYLA and BOBYQA solvers are non-derivative-based, while MMA, CCSAQ, and SLSQP, and PANOC solvers are derivative-based.  All gradients for the derivative-based solvers were calculated using finite differencing.  We compared the performance single-threaded versions of these algorithms to versions that use the Strobe acceleration meta-algorithm on 12 threads.  All paths in this experiment contained 100 waypoints.  

Our results, summarized in Figure \ref{fig:experiment1}, show that the Strobe algorithm greatly accelerates the path optimization process, often by over an order of magnitude.  Also, the objective metrics indicate that this performance gain did not come at the expense of optimization quality.  In fact, Strobe often achieved greater optimization quality, which we speculate may be because it is somewhat rare for all of the pods to fall into local minima at the same time.

\subsection{Experiment 2: Varying Number of Threads}

In Experiment 2, we vary the number of threads to assess how Strobe scales with more parallelization.  We only used the SLSQP optimization algorithm in Experiment 2, and paths in this experiment contained 100 waypoints. 

Our results for Experiment 2 can be seen in Figure
\ref{fig:experiment2}.  We see that Strobe scales successfully to more threads, again without sacrificing optimization quality.    

\subsection{Experiment 3: Varying Parallelization Scheme}
\begin{figure}[t!]
	\includegraphics[width=\columnwidth]{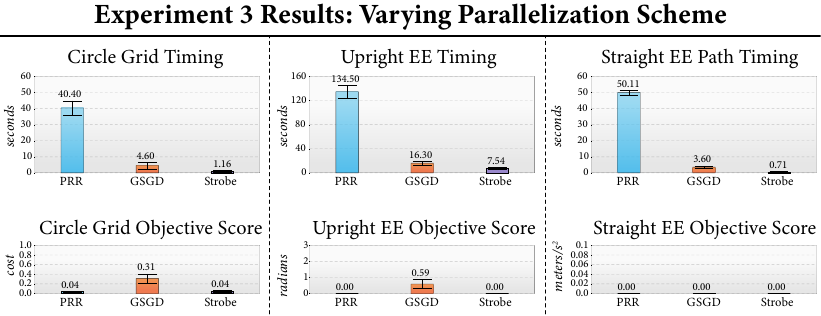}
	\caption{Results from Experiment 3.  Range values denote standard error.  }
	\label{fig:experiment3}
	\vspace{4pt}
\end{figure}

In Experiment 3, we compare Strobe to other possible parallelization schemes.  Specifically, we compare against \textit{parallel random restart} (PRR) that uses all available threads to try optimizing from different initial conditions.  The algorithm stops as soon as one thread successfully converges with respect to a given tolerance.  We also compare against \textit{generalized stochastic gradient descent} (GSGD) that is meant to resemble the common SGD strategy, but applied to other optimization algorithms other than just gradient descent.  This condition involved each thread randomly selecting a subset of the waypoint path and optimizing over this subset.  We again only used the SLSQP optimization algorithm in Experiment 3, and paths in this experiment contained 100 waypoints.  

Our results from Experiment 3 can be seen in Figure \ref{fig:experiment3}.  We see that Strobe outperforms the alternative parallelization schemes in terms of both run-time and optimization quality.  GSGD did not find optimal solutions for all tasks.  We believe this condition had a difficult time successfully converging because parallel optimizations sometimes overlapped and inadvertently disrupted and contradicted the optimization strategies of threads nearby.  This appears to be a particular problem when optimizing over paths due to the structured and ordered nature of the problem.  In contrast, Strobe ensures that none of its individual optimizations are affecting any other optimizations nearby, and we find that it successfully converges on our evaluation tasks.   
\subsection{Experiment 4: Varying Number of Waypoints}

In Experiment 4, we vary the number of waypoints along the path being optimized.  We compare the performance of single-threaded SLSQP with SLSQP using Strobe on waypoint paths of length 25, 50, 100, and 200.  

\begin{figure}[t!]
	\includegraphics[width=\columnwidth]{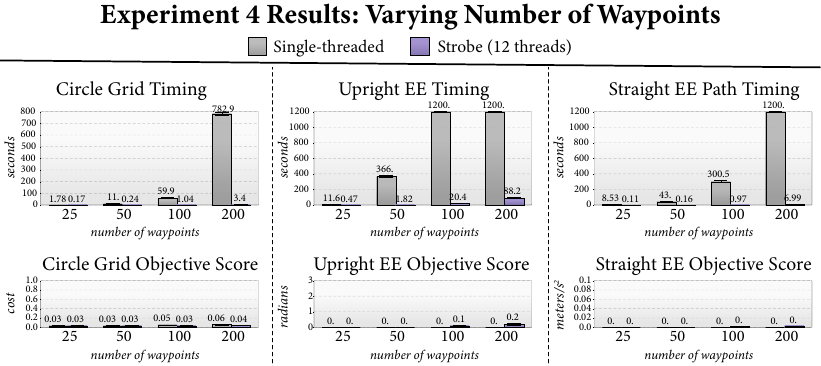}
	\caption{Results from Experiment 4.  Range values denote standard error.  Gray bars on the left show results for the optimization algorithm run on a single thread, while the purple bars on the right show the results for the optimization algorithm run using the Strobe meta-algorithm.}
	\label{fig:experiment4}
	\vspace{4pt}
\end{figure}

Our results from Experiment 4 can be seen in Figure \ref{fig:experiment4}.  We see that Strobe scales significantly better on denser paths than its single-threaded counterpart.  In fact, SLSQP on a single thread did not find a solution within the maximum time (20 minutes) on the \textit{Upright EE} or \textit{Straight EE Path Timing} when using 200 waypoints.  In contrast, these computations remain tractable using Strobe.  On the sparser paths, we see that Strobe also greatly accelerated the computations.  Often, the path successfully converged in tens of milliseconds.  

\section{Discussion}
\label{sec:discussion}
In this work, we presented an acceleration meta-algorithm for robot path optimization based on the premise of splitting the input path into separate pods, and optimizing over various pods concurrently.  In this section, we discuss the limitations and implications of our work.  

\subsection{Limitations}
We note a number of limitations of our work that suggest future extensions.  First, despite the connections between our work and trajectory optimizers, we have not shown our method to work on full trajectories that optimize over state and control variables over time.  We plan to explore this exciting direction in future work.  Second, while our pod splitting strategy appears to be successful across many scenarios, even compared to other parallelization schemes, we cannot claim that this is an optimal way to split and optimize over path subsets.  Extensions to this work could characterize the speed and quality of convergence based on particular path splitting strategies.  We also note that a theoretical analysis of our meta-algorithm was beyond the scope of our current work.  In particular, the convergence behavior at the boundaries between blue and red pods may be of particular interest for future theoretical investigations.  Further, we cannot guarantee that our method will enable a solver to converge to the same solution that it would have if the solver considered the whole path.  Lastly, while showing more optimization scenarios and more optimization algorithms was beyond the scope of this work, we believe characterizing our method across more domains will be important in extensions to this work.     

\subsection{Implications}
At a high level, we expect our method to have three levels of practical benefit: (1) Relatively sparse paths could be optimized in tens of milliseconds, meaning optimization at this level would be amenable for many real-time, performance critical control loops; (2) Moderately dense paths could be optimized in hundreds of milliseconds, meaning optimization over reasonably rich paths would be usable in many motion refinement loops in real-time settings; and (3) dense paths could be optimized in a few seconds, meaning optimization over complex and highly parameterized paths would be practical in semi-online, interactive applications.  We believe these benefits could impact many areas of robotics, such as teleoperation, shared-control, or model predictive control.  



\balance
\bibliographystyle{IEEEtran}
\bibliography{references}

\end{document}